\documentclass[a4paper, 10pt, conference]{IEEEconf} 
\IEEEoverridecommandlockouts                              

\usepackage{hyperref}
\usepackage{cite}
\usepackage{amsfonts}
\usepackage{algorithmic}
\usepackage{graphicx}
\usepackage{textcomp}
\usepackage{setspace}
\usepackage{amsmath}
\usepackage{amssymb}

\usepackage{amsthm}
\usepackage{graphicx}
\usepackage{mathtools}

\usepackage{colortbl}
\usepackage{caption}
\usepackage{subcaption}
\usepackage{balance}
\usepackage[linesnumbered,ruled,vlined]{algorithm2e}

\usepackage{siunitx}
\usepackage{hyperref}

\usepackage[most]{tcolorbox}
\usepackage{float}
\usepackage{bm}

\title{Accelerated gradient descent for high frequency Model Predictive Control}
\author{Jianghan Zhang$^{1}$, Armand Jordana$^{1}$ and Ludovic Righetti$^{1}$
\thanks{This work was in part supported by the National Science Foundation grants 1932187, 2026479, 2222815 and 2315396.}
\thanks{\textsuperscript{1}~Machines in Motion Laboratory, New York University, USA
        {\tt\small jz5480@nyu.edu aj2988@nyu.edu, lr114@nyu.edu}}}

\begin{document}
\maketitle

\begin{abstract}
The recent promises of Model Predictive Control in robotics have motivated the development of tailored second-order methods to solve optimal control problems efficiently. While those methods benefit from strong convergence properties, tailored efficient implementations are challenging to derive. In this work, we study the potential effectiveness of first-order methods and show on a torque controlled manipulator that they can equal the performances of second-order methods. 
\end{abstract}

Model Predictive Control (MPC) is a potent framework for online control of robots. It has showcased outstanding performance on different kinds of robots, ranging from manipulators\cite{9560990}, quadrupeds\cite{9387121},\cite{meduri2023biconmp}, to humanoids\cite{dantec2022whole}. MPC formulates the control of robots as the solution to a nonlinear optimization problem. More precisely, at every control cycle, an optimal control problem (OCP) is solved online using the current state of the system as the initial state, and the first control input is applied to the system. 
This online re-planning enables robots to adjust to external disturbances and changes in their environment.

Historically, the robotics community has favored second-order methods like Differential Dynamic Programming (DDP)\cite{mastalli20crocoddyl} or Sequential Quadratic Programming \cite{jordana:hal-04330251} to efficiently solve OCP online. Although second-order methods allow to obtain high accuracy solutions efficiently, they require second-order information on the cost and dynamics, which are tedious to derive and implement. In particular, there have been decades of research on analytical derivatives of rigid body dynamics\cite{10.5555/1324846}, and much effort has been invested in the efficient derivation and implementation of first-order derivatives~\cite{8700380}. More recently, \cite{10449483} provided an efficient way to compute the second-order terms of the dynamics, and \cite{singh2023analytical} showed how to use them with DDP. Nevertheless, in order to use second-order methods, practitioners still need to implement the second-order derivatives of each cost function term used, which can be tedious in practice. Consequently, popular software such as~\cite{mastalli20crocoddyl}, use Gauss-Newton approximation to avoid the computation of second-order terms. Moreover, second-order methods require regularization and line search schemes in the non-convex case \cite{NoceWrig06}. 
Lastly, efficient implementation need to exploit the time-induced sparsity of the structure, which needs a significant amount of implementation effort\cite{wright1990solution, jordana:hal-04330251}. 

In contrast, first-order methods only need access to first-order information on the cost and dynamics. They are much simpler to implement and computationally cheaper while having the potential to provide reasonable approximate solutions by performing more iterations than second-order methods in the same amount of time.
Moreover, for MPC, approximate solutions of OCP might be enough to control robots \cite{meduri2023biconmp}. Hence, one might ask: \textbf{Can first-order methods achieve the same performance as second-order methods when used at high frequency in MPC}?

\begin{figure*}[h!]
        \centering
        \includegraphics[width=0.92\textwidth]{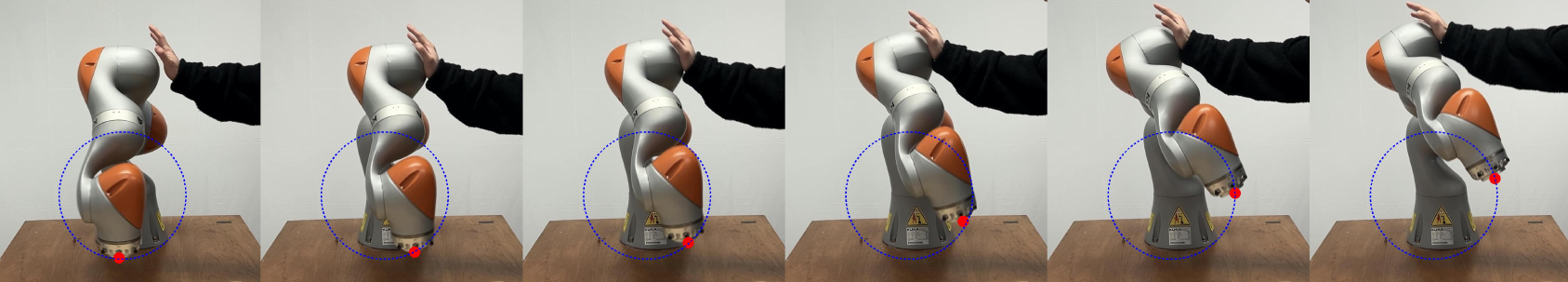}
        \caption{Circle tracking in the end effector space under external disturbance}
        \label{fig:disturbance}
\end{figure*}

First-order methods are a well-established optimization technique which have been studied for decades\cite{beck2017first} and have extensive applications in fields like machine learning\cite{lan2020first}. There has been some interest in the control community for first-order methods.  In \cite{5531095}, Nesterov's accelerated gradient method is used to solve a linear MPC problem directly. In\cite{8263933}, a quasi-Newton method with an forward-backward envelope based line-search is used to solve the OCP, which was then applied to solve obstacle avoidance problems in MPC\cite{Sathya_2018}. Nevertheless, to the best of our knowledge, first-order methods have not been used to perform high frequency MPC on robots.

In this work, we study the feasibility of using accelerated gradient descent (AGD) for nonlinear MPC on a real robot. The employed algorithm is an instance of Gradient Descent, integrated with an Adaptive Momentum method (ADAM) for faster convergence\cite{kingma2014adam}. The algorithm exhibits a linear complexity in the time horizon and the dimensions of state and control. The algorithm uses a single shooting approach, which considers the control sequence as decision variables. The gradient is calculated with a backward pass similar to DDP using only first-order information. The algorithm has no matrix-matrix multiplication nor matrix inversion. 
Given the update on the control sequence, a forward pass with a nonlinear roll-out is performed to obtain the state sequence update, which enforces dynamic feasibility. In the MPC setting, both the initial guess and the momentum terms of ADAM are warm-started with the solution from the previous time step.
The adaptive nature of ADAM makes it possible to avoid the use of a line-search.
This reduces computational costs per iteration, enabling more iterations within each control cycle. More specifically, our method can perform an iteration four times faster than DDP. 

\begin{figure}[ht]
        \centering
        \includegraphics[width=0.48\textwidth]{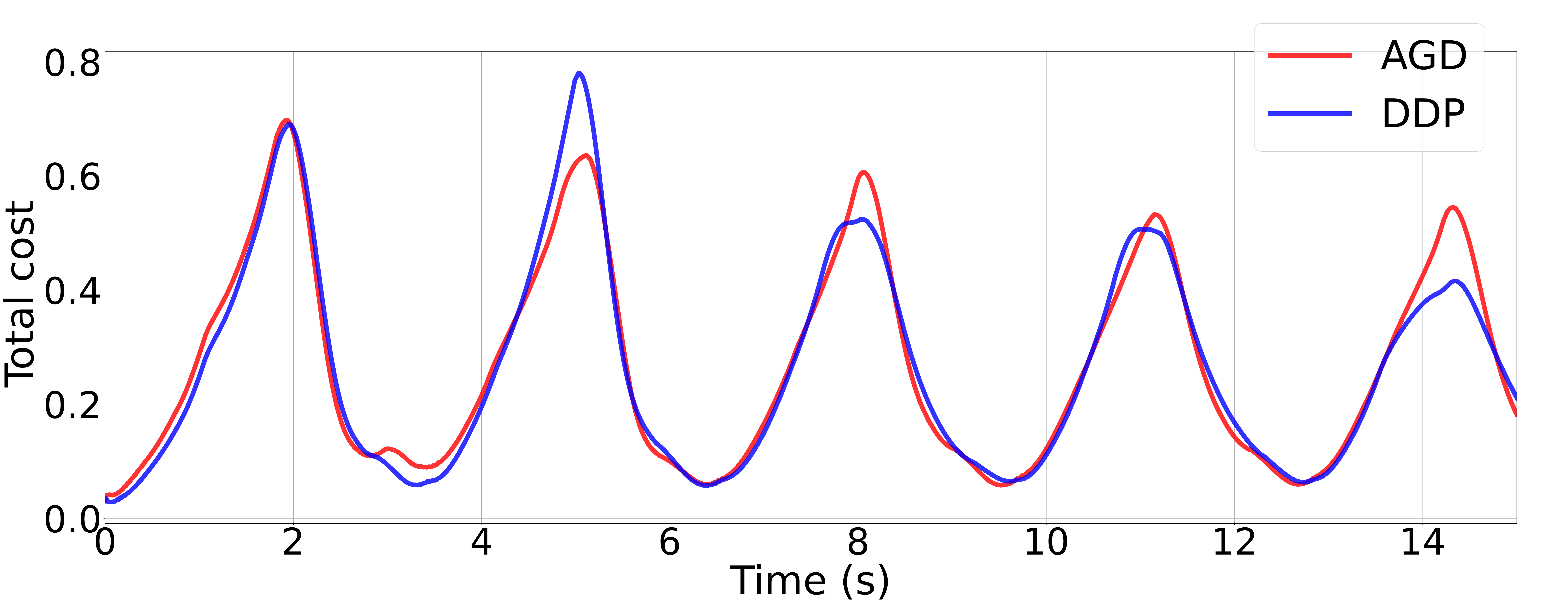}
        \caption{Comparison of the running cost for the first-order (AGD) and second-order method (DDP)}
        \label{fig:cost}
\end{figure}

\begin{figure}[ht]
        \centering
        \includegraphics[width=0.48\textwidth]{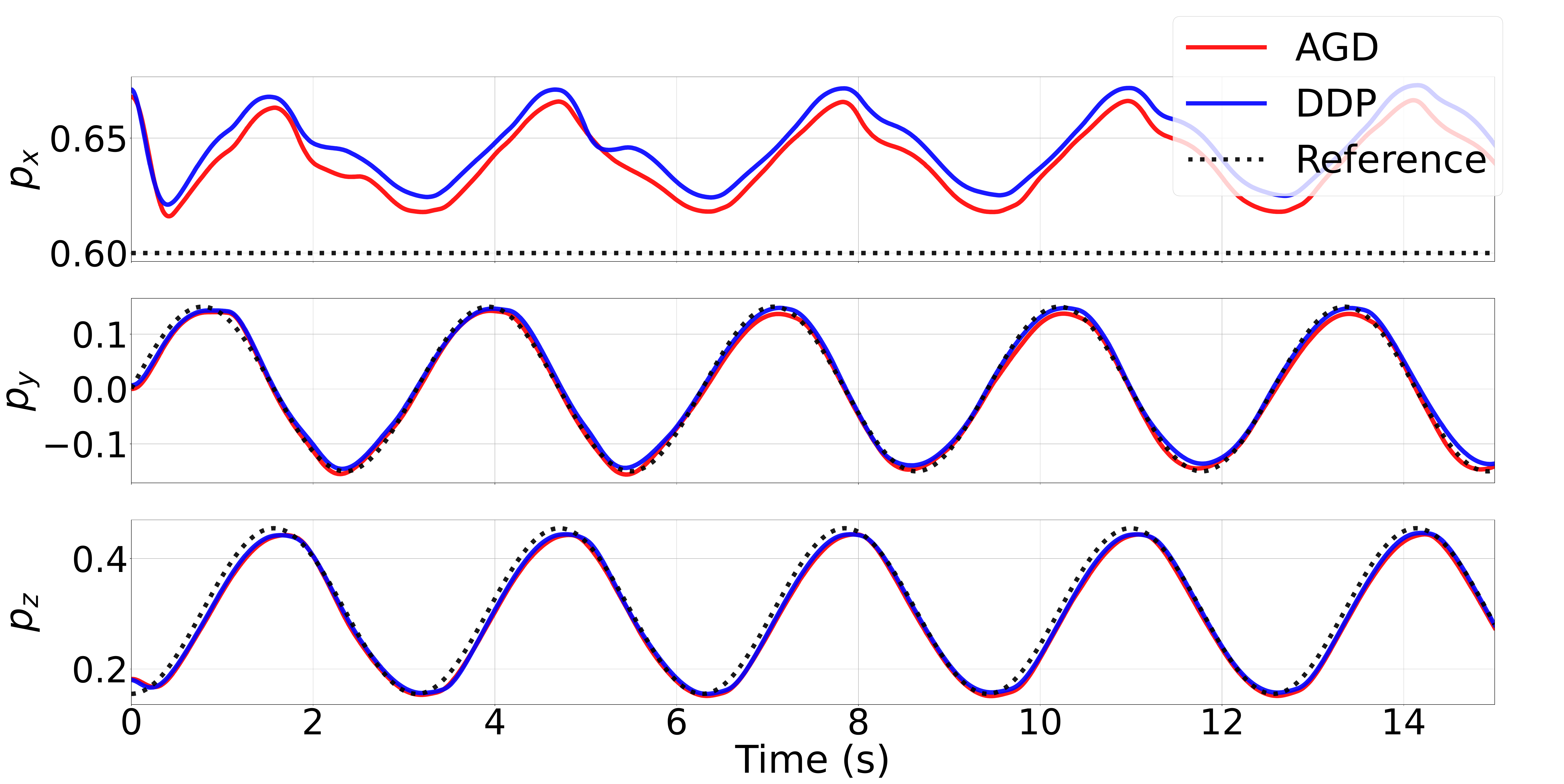}
        \caption{End-effector positions of the robot for the first-order (AGD) and second-order method (DDP)}
        \label{fig:position}
\end{figure}

We demonstrate the effectiveness of this first-order method on a 7 degree of freedom torque-controlled manipulator and compare it against a DDP. More precisely, we demonstrate that we can perform 1KHz nonlinear MPC with AGD without any performance loss compared to DDP. To the best of our knowledge, this is the first demonstration of using a first-order method to solve high-frequency closed-loop nonlinear MPC on real hardware.
The robotic setup follows our previous work~\cite{9560990}, where more details are available. The cost function includes state and torque regularization, as well as an end-effector position tracking term. For MPC, we control the robot at 1kHz, and for every millisecond, the AGD makes 8 iterations while DDP makes 2. In a first experiment, we compare AGD and DDP in a circle tracking task in end-effector space. Figures \ref{fig:cost} and \ref{fig:position} show that from both the running cost of MPC and the circle tracking error. AGD achieves comparable performance with DDP. In a second experiment, we study the robustness of AGD on the same circle tracking task. We exerted external disturbances from a human operator and observed that AGD could ensure a compliant response to external disturbances while performing the designated task (see Figure \ref{fig:disturbance}). These two experiments suggest that the first-order methods can achieve comparable performance with second-order methods. Our code is open-source and available online\footnote{\scriptsize{\url{https://github.com/JianghanZHang/AGD}}}. The possible future works include imposing general inequality constraints by using, for example, Augmented Lagrangian methods. 
\vspace{-0.5cm}

\bibliographystyle{IEEEtran}
\balance
\bibliography{references}

\end{document}